# Method and Software Tool for Generating Artificial Databases of Biomedical Images Based on Deep Neural Networks

Oleh Berezsky*a*, Petro Liashchynskyi*a*, Oleh Pitsun*a* and Grygoriy Melnyk*a*

*a West Ukrainian National University, 11 Lvivska st., Ternopil, 46001, Ukraine*

#### Abstract
A wide variety of biomedical image data, as well as methods for generating training images using basic deep neural networks, were analyzed. Additionally, all platforms for creating images were analyzed, considering their characteristics. The article develops a method for generating artificial biomedical images based on GAN. GAN architecture has been developed for biomedical image synthesis. The data foundation and module for generating training images were designed and implemented in a software system. A comparison of the generated image database with known databases was made.

#### Keywords 1
Breast cancer, image generation, generative adversarial networks. training data sets, digital platform, artificial databases of biomedical images

## 1. Introduction

Currently, deep neural networks (DNNs) are widely used for automatic diagnosis in medicine. Training data sets (TND) are used to train DNNs. Training datasets help improve the accuracy of automatic diagnosis. The following biomedical images are used to make a diagnosis in oncology: cytological, histological and immunohistochemical images. Original sets of these images are limited. This is explained by the objective reasons for their receipt.

Sources for creating data sets include both international scientific projects and competitions in the development and testing of algorithms (for example, Kaggle). To distribute data sets, resources from individual projects, conferences and competitions (challenges), as well as special platforms are used. Examples of such special platforms for publishing datasets are: Data.gov[1], Kaggle Datasets[2], Zenodo[3], Google Dataset Search[4] and many others.

When designing biomedical image analysis systems, the problem of increasing TND is also relevant. Increasing the TND helps improve the accuracy of models and takes into account rare classes. To increase the data set, you can use augmentation and synthesis tools.

Augmentation is the process of artificially increasing the size of a TND by using various image transformations. These can be operations such as shifting, scaling, rotating, changing contrast and others, allowing you to get different variations of the same image.

Synthetic data is created artificially using an initial small dataset. Generative adversarial networks (GANs) are used to create synthetic data.

Purpose of artificial (synthetic) image sets:

1. Training and testing of algorithms. Synthetic datasets allow developers and researchers to create images of known pathologies. This is useful when real-world clinical data are limited or unavailable.

2. Studying the stability of algorithms to various artifacts that may appear in real images.

The characteristics of artificial (synthetic) image sets are as follows:

1. Synthetic datasets allow you to control image parameters, such as resolution, types of pathologies, degree of complexity, and others. This facilitates the study of specific aspects of diagnosis.

2. When creating synthetic images, the true state of the pathology is known, which allows you to accurately evaluate the effectiveness of the algorithms and analyze their accuracy.

3. Using synthetic data, large volumes of images can be generated for testing and validating algorithms at different scales.

4. Public access makes these and other datasets available to the public, facilitating data sharing and research collaboration.

Therefore, an actual problem is the generation of biomedical images in oncology. This provides the necessary accuracy in the classification of biomedical images. To solve this problem, GAN was used in this work.

## 2. Literature review

In their fundamental work, Ian Goodfellow and others developed the concept of Generative Adversarial Networks (GANs) [5]. The developed architecture consists of two neural networks – a generator and a discriminator. The generator produces data for the discriminator. The discriminator distinguishes genuine data from generated data. However, the disadvantages of such networks remain potential training instability and non-convergence of the model.

In work [6], the potential of GANs for datasets focused on liver pathologies was explored. The main theme of the research was the ability of GANs to combat the common problem of overfitting in deep learning models. The paper demonstrated the improvement and extension of medical data sets, which increased the reliability of the model.

Synthesizing medical images using GANs involves not only extension data sets, but also ensuring the confidentiality of patient information. Work [7] used GAN to create synthetic medical data to ensure patient data privacy by adding noise to images.

The authors of article [8] used GAN to create MRI images of the heart. The goal of the work is to study the influence of architectural features of networks on the quality of synthesized images. Ensuring high resolution and accuracy of synthesized images are major problems.

Other research [9,10] has shown that GANs can be used to synthesize images of retinal and lung cancer cells. The synthesized lung cancer nodules passed the visual Turing test with the participation of radiologists.

The article [11] explored the problem of "model collapse," when the generator synthesizes a small number of images, thus limiting the variety of synthesized images. The article suggests that the difficulty of training GANs and the need for large amounts of training data limit the application of GANs in medicine.

The authors of work [12] showed that a GAN network is also capable of learning to simulate the distribution of all high-resolution MRI images.

To synthesize high-resolution images of skin lesions, the researchers compared several GAN network architectures. A classifier was successfully trained based on the synthesized samples [13]. Using the concept of progressive growth of GANs, the authors of [14,15] generated realistic synthetic images of skin lesions.

The method has been developed in [16] that allow synthesizing histopathological images. Based on specific tissue types, cancer subtypes, and known reference image data, the authors synthesize histopathological images.

The authors of the article [17] used an improved conditional GAN architecture, which they called HistoGAN. The researchers used the self-attention module and other methods to stabilize the learning process and improve the quality of the synthesized images.

The authors of the article [18] propose a generative competitive network with additional regularization based on loss of sharpness to generate realistic histopathological images. The authors additionally introduce a sharpening loss that enhances the contrast of pixels on the contours of the nuclei.

Researchers in [19] propose a vision transformer-based GAN model for synthetically augmenting a set of histopathological images.

The authors of this work have developed a many of CADs for automatic diagnosis in oncology [20-24].

Therefore, GANs have significant potential in biomedical image synthesis. However, the application of GANs is influenced by training complexity, data quality, ethical issues, and clinical relevance.

## 3. Problem statement

The conducted analysis showed that there is a problem of increasing datasets for the classification of biomedical images. To solve of this problem, it is necessary to:
- analyze the typical training sets of images;
- develop the artificial image synthesis method;
- develop the software synthesis and image storage;
- conduct the computer experiments.

## 4. Analysis of typical training sets of images

World practice has shown that the creation and storage of data sets is a pressing problem. Creating datasets is a separate task. A large number of specialists from different countries are involved in the creation of data sets. Data sets are stored on well-known platforms. This is also a separate task. Therefore, we will first analyze the training data sets.

The main characteristics TND [25] are as follows:
1. Image volume – the number of images of normal and pathological tissues of the human body.
2. Image format. The most efficient way to present images is through WSI (Whole Slide Imaging) multi-scale whole slide imaging. To train algorithms, TND must contain instructions or expert segmentation of micro-objects. This segmentation is maintained through contour coordinates and binary masks.

Automated microscopy systems are used to create TND. For example, the Automated Slide Analysis Platform (ASAP) is an open platform for visualization, annotation, and automated analysis of WSI images. ASAP is built on the basis of well-known open source software OpenSlide, Qt and OpenCV.

Camelyon16, Tupac16 are TDSs playing an important role in the development of modern digital histopathology and cancer diagnosis. Based on large volumes of tissue imaging, these kits provide researchers and clinicians with a unique opportunity to study and develop advanced machine learning algorithms for automated analysis of histological data.

The Camelyon16 dataset [26] was built to detect lymph node metastases for breast cancer. It includes about 400 gigabytes of lymphatic images, which are used to train and test the algorithms.

The TUPAC16 dataset [27, 28] was created to investigate the morphological characteristics and detect cancerous abnormalities in the mammary glands. The set includes 490 gigabytes of images. The Camelyon17 dataset is an extended version of Camelyon16, containing even more images and clinical data.

Platforms are formed based on data sets. Let's analyze the main functions of the platforms. The main functions of the platforms are:
1. Data storage and organization. Platforms allow you to load, store and organize data in hierarchical structures. Versioning, GitHub integration, visit and download statistics can also be supported.
2. Availability and exchange. Data can be shared with the global community.
3. Methods and description of data sets. The platforms allow you to provide detailed descriptions and metadata for each dataset. Metadata includes DOI, author information (ORCID), keywords, project and research grant information, source citations, organization type, publisher type.
4. Licenses and access control. Users can set licenses and rules for accessing datasets. The following are available for public sets: Creative Commons, GPL, Open Data Commons, Community Data License. Collaboration functionality is also available, allowing multiple users to jointly own and maintain a private or publicly accessible dataset.
5. Tools for analysis and rendering. Some platforms provide tools for analyzing, processing and visualizing data directly on the platform.

6. Support for data formats. Platforms can support different data formats, including text, images, video, audio and others. In particular, the available formats are XML, PDF, HTML, EXCEL, CSV, JSON, RDF, DOC, ZIP, BigQuery.

Comparative characteristics of known platforms are presented in Table 1.

## 5. Artificial image synthesis method

A method for synthesizing artificial biomedical databases using GANs has been developed in this research.

GAN is a neural network consisting of two neural networks: a generator and a discriminator, trained simultaneously using a competitive process. The generator tries to create realistic data, the discriminator distinguishes genuine data from artificially generated ones. Over time, the generator improves its ability to create realistic data by learning from feedback from the discriminator.

**Table 1**
Comparative characteristics of platforms

| Characteristics/ Platform | Data.gov [1] | Kaggle Datasets [2] | Zenodo [3] | Google Dataset Search [4] |
|---|---|---|---|---|
| Type | Government resource | Commercial platform | Academic repository | Search tool |
| Openness | Open data | Open and closed data | Open data (sometimes with restrictions) | Mostly open data |
| Categories of data | Government data (education, health, economy, etc.) | Diverse (from science to entertainment) | Academic and research data | Various (from various sources) |
| File formats | CSV, JSON, XML, etc. | CSV, SQLite, JSON and others | PDF, CSV, ZIP, other scientific formats | Depends on the source |
| API | Yes | Yes | Yes | No |
| Additional functions | Resources for developers, visualization tools | Competition, cores (code), discussions | Digital DOIs, integration with GitHub | Search by metadata |

The developed method consists of the following steps (Figure 1):
1. Loading training images from the directory.
2. Extension of the training sample by applying affine distortions. In the developed method, the following affine distortions are applied to images: random scaling, rotation, shift. All operations are used with a 50% probability. The software implementation of distortions is based on the Rudi library [29].
3. Train and evaluate the GAN network using the extended training set from the previous step. The network is trained for a given number of iterations.

To assess the quality of generated images, the following metrics are used: *FID* (Frechet Inception Distance) and *IS* (Inception Score). Both metrics are based on the Google Inception 3 classifier model, designed for classifying color images and trained on the ImageNet dataset. The FID metric helps evaluate the quality of the generated images, and the IS metric helps evaluate the diversity of images. Smaller FID values indicate better quality of the synthesized images, and larger IS values indicate better diversity. The model is evaluated based on metrics every *M* training iterations.

Once a given number of *Iter* iterations is reached, we generate *N* images and store them and the trained model in a database.

Thus, our method combines data loading, data augmentation, GAN training, image generation, and image quality evaluation.

To generate the image, use the GAN. The architecture of the generator and discriminator is based on ResNet Block. Also in the generator and discriminator there is a self-attention mechanism. The architecture of GenBlock and the generator is shown in Figures 2 and 3.

The architecture of the DiscBlock and discriminator is shown in Figures 4 and 5.

The following training parameters were used during generation:
1. Optimizer – Adam.
2. Generator learning rate 1e-4.
3. Discriminator learning rate 4e-4.
4. Loss function – Hinge Loss.
5. Epochs – 100,000.
6. Batch size – 96.

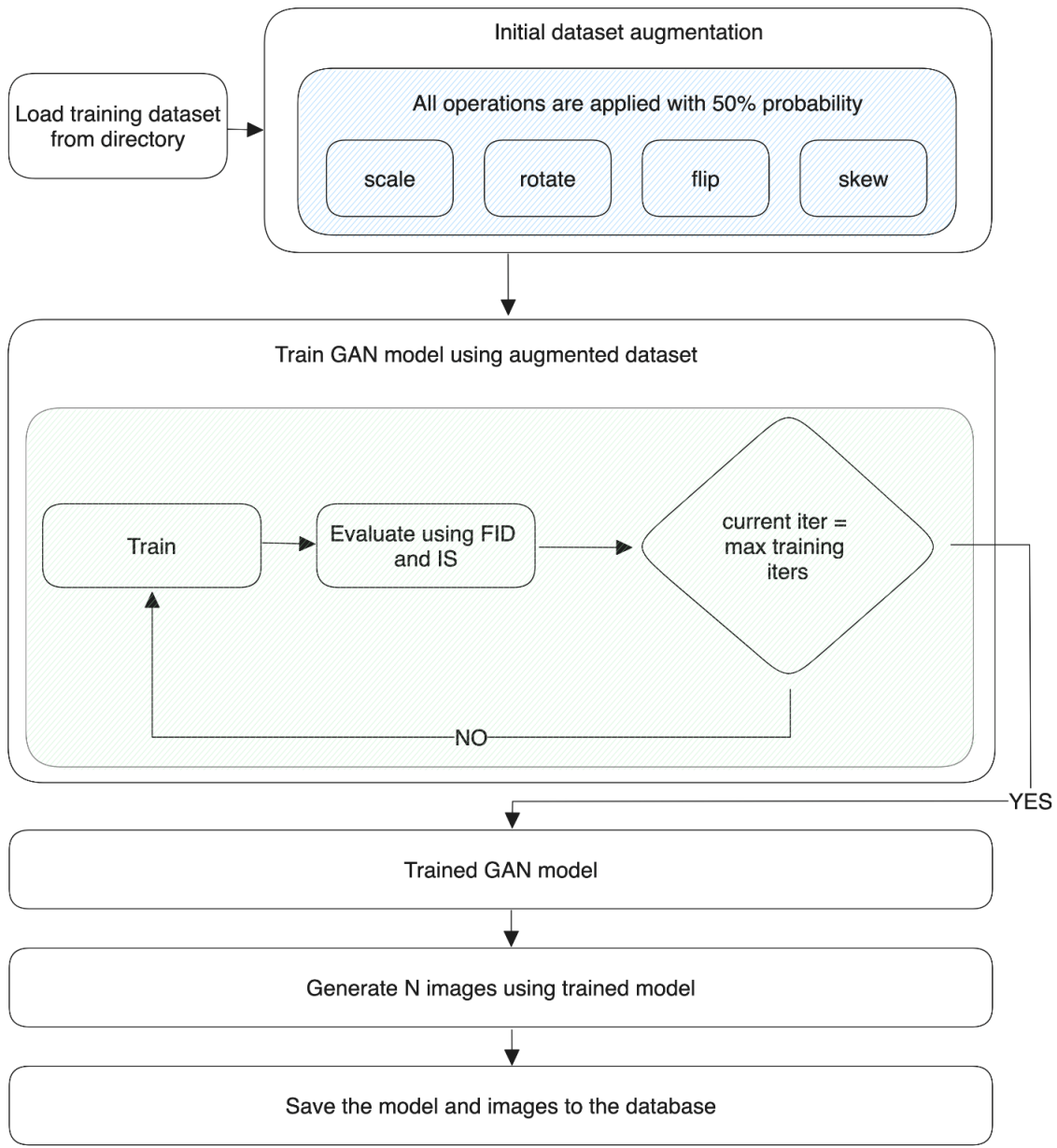

**Figure 1:** The method of image generation and storage

# 6. Software synthesis and image storage

The implementation of the software is based on the use of the Google Cloud Platform (GCP) cloud infrastructure. This approach allows efficient use of data storage resources. The program infrastructure is shown in Figure 6.

The software implementation has two main files: train.py and generate.py. The train.py file is for training a PyTorch GAN model. This file defines the architecture and parameters of the GAN model. To train the model, we use the Vertex AI service, which provides infrastructure and tools for efficient model training (with GPU support).

The FID and IS metrics are used to evaluate the performance of the trained model.

After graduation, the model is uploaded to Cloud Storage. Next, the trained GAN model is deployed on the Vertex AI platform by generating a final URL. This address can be accessed to create images. The deployment process involves specifying the necessary computing resources and loading our model code into a Docker container for efficient use.

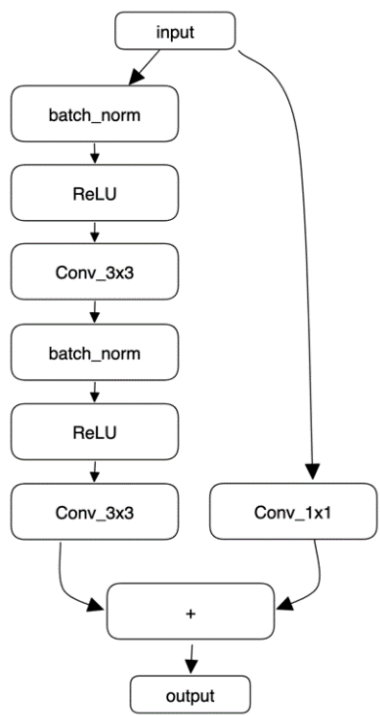

**Figure 2:** GenBlock

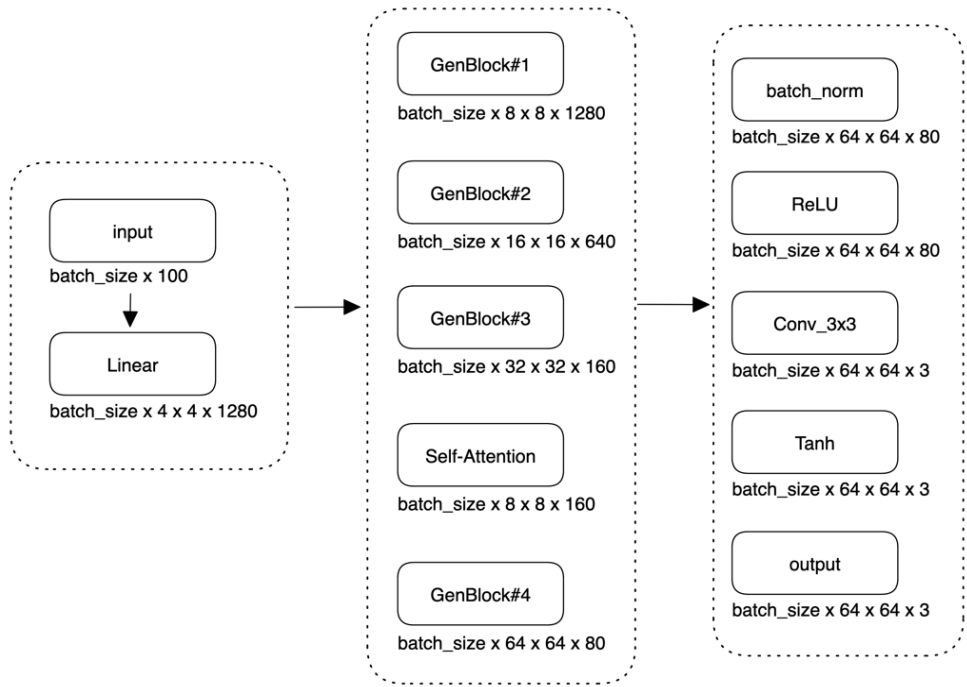

**Figure 3**: Generator

Once the model is deployed, the master data is stored in a Cloud SQL instance. This database is a central repository for storing the following data: Model endpoint URLs, Image scoring metrics. This allows you to track model versions and associated performance characteristics.

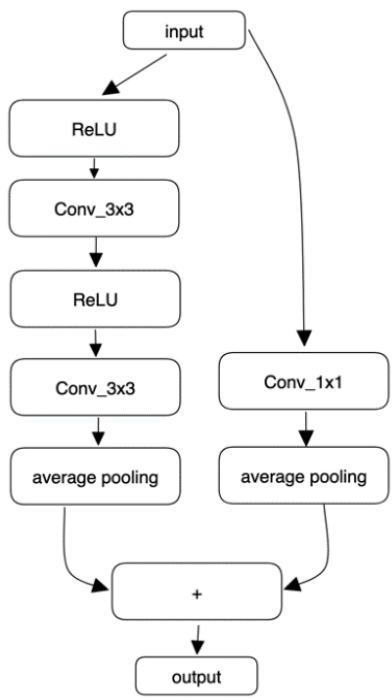

**Figure 4:** DiscBlock

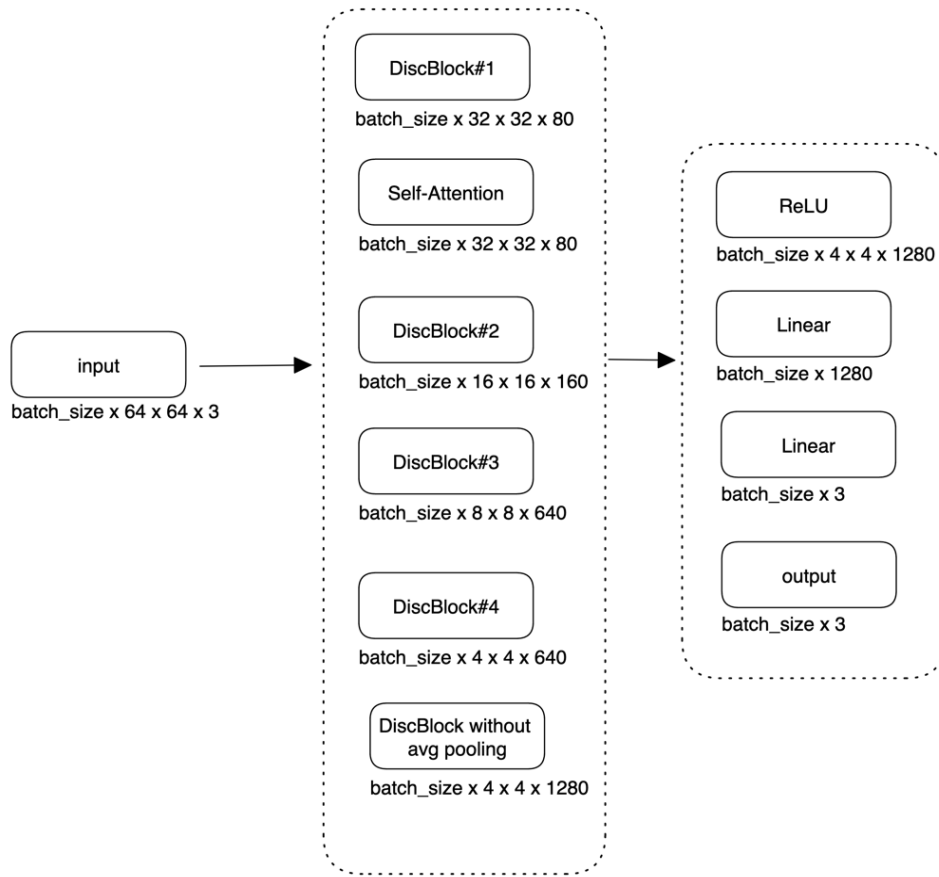

**Figure 5**: Discriminator

The generate.py file is used to create new images. Users can specify the number of images to create. An optional parameter is the identifier of the model used for image synthesis. By default, the last model added to the database is used. For example, to generate 1000 images, we use model ID 5. To do this, just run the program by calling python generate.py 1000. 5. This script interacts with the URL endpoint of the deployed model, sending requests to generate images.

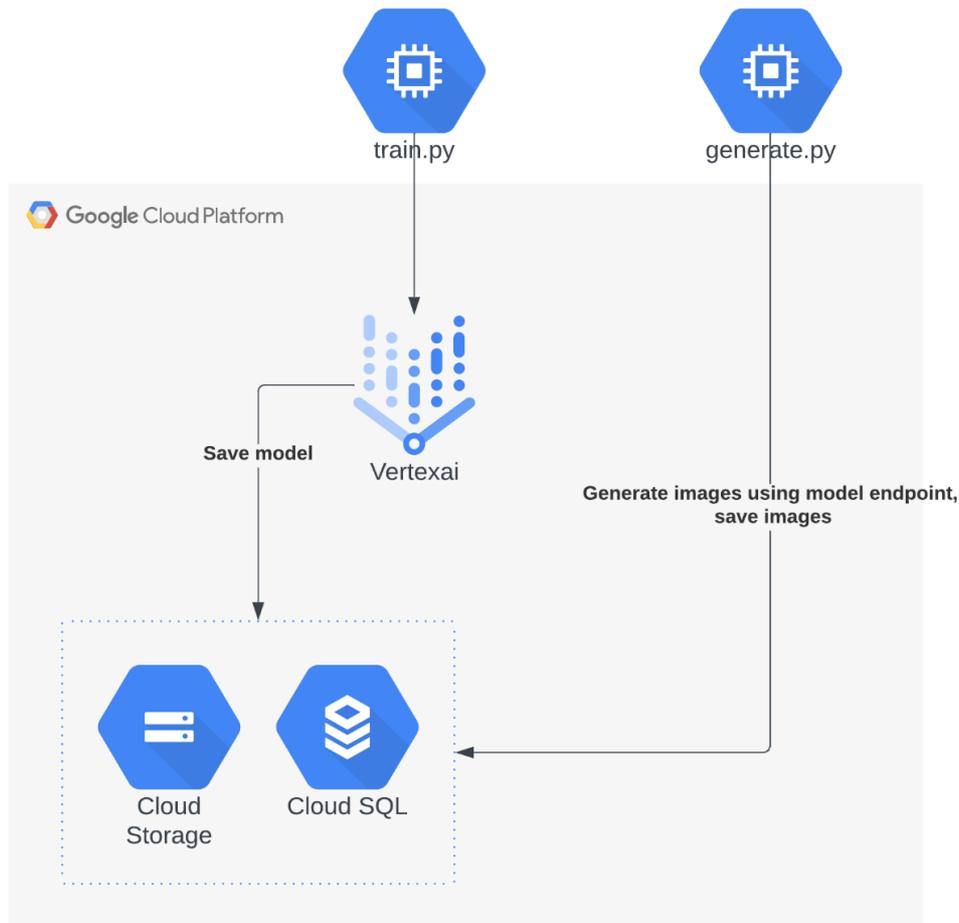

**Figure 6:** Cloud infrastructure of the program

The generated images are obtained and stored in GCP cloud storage. At the same time, information about the created images is written to the cloud SQL instance of PostgreSQL. The database schema is shown in Figure 7.

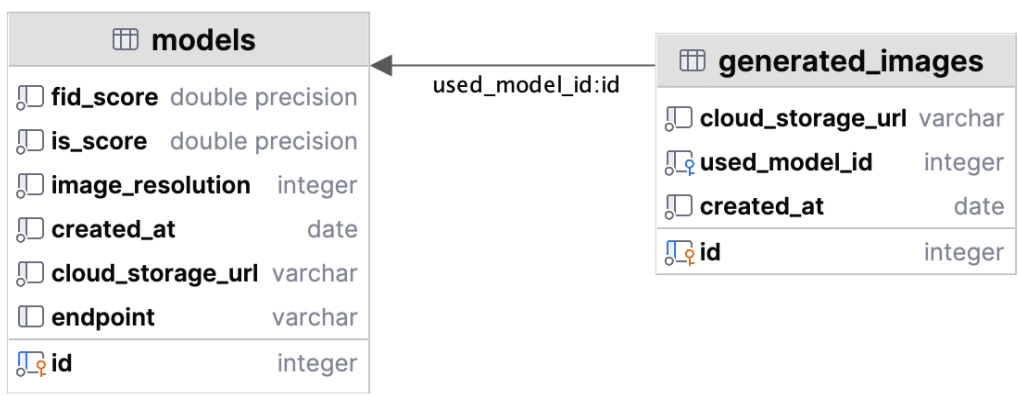

**Figure 7**: PostgreSQL database schema

Thus, our software infrastructure integrates Python, PyTorch, Google Cloud Platform services such as Vertex AI, Cloud Storage, and PostgreSQL Cloud to facilitate GAN model training, deployment, and management. The generate.py script provides a convenient interface for creating new images while maintaining a record of the model's performance and generated content in cloud storage and a database. This architecture makes it possible to efficiently generate images for various purposes.

## 7. Computer experiments

A dataset was used to conduct computer experiments. The dataset is a sample of histological images measuring 64 by 64 pixels. The dataset is divided into three classes. The total number of images is 185. This sample is expanded to approximately 700 images by applying a set of affine transformations (random rotation, translation, scaling).

An example of the start date of the set is given in Table 2.

**Table 2**
Characteristics of the initial dataset

| Type of images | Color RGB histological images | | |
|---|---|---|---|
| Image classes | G1 | G2 | G3 |
| Example images for each class | 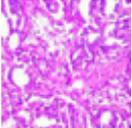 | 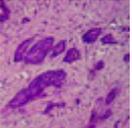 | 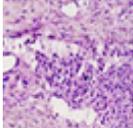 |
| resolution | 64 by 64 pixels | 64 by 64 pixels | 64 by 64 pixels |
| The total number of images in the dataset | | 700 | |

For the experiments we used the GCP n1-standard-4 virtual machine: 15 GB RAM, 4 vCPU, Nvidia Tesla V100 GPU 16 GB (13.2 TFLOPS). The network training process lasted about 11 hours. The Inception Score (IS) and FID metrics were used to evaluate the network. The metrics values are as follows: IS – 3.025, FID – 68. As a result of the experiments, 2000 artificial images were generated for each class. The resolution of the generated images is 64 by 64 pixels. An example of these images is shown in Figure 8.

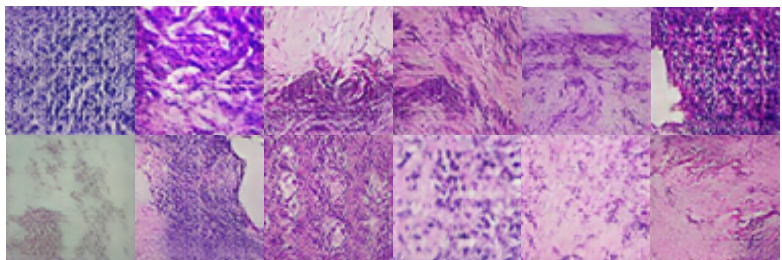

**Figure 8:** Example of generated images

The results of the experiments are stored in the Cloud SQL database. Examples of records from the database are shown in Figure 9.

| id | cloud_storage_url | used_model_id | created_at |
|---|---|---|---|
| 1 | /gcs/experiments/gen_images/1.png | 1 | 2023-09-20 |
| 2 | /gcs/experiments/gen_images/2.png | 1 | 2023-09-20 |
| 3 | /gcs/experiments/gen_images/3.png | 1 | 2023-09-20 |
| 4 | /gcs/experiments/gen_images/4.png | 1 | 2023-09-20 |
| 5 | /gcs/experiments/gen_images/5.png | 1 | 2023-09-20 |

**Figure 9**: Application records from tables of generated images

## 8. Conclusions

1. The known and available training datasets were analyzed, revealing the limitations of rare classes in biomedical dataset images.
2. Global image storage platforms were analyzed, their characteristic functions were highlighted. The conducted analysis showed that the number of artificial databases of biomedical images is small.
3. The method for generating artificial images has been developed, which consists of the following steps: affine distortions of the original images, generation of images based on GAN, and evaluation of the quality of the generated images.
4. The GAN architecture, which consists of a generator and a discriminator, has been developed. The basis of the discriminator and generator is the ResNet Block. The self-attention mechanism is used for the generator and discriminator.
5. Software module for generating and storing artificial images in the Python programming language was developed. Software infrastructure combines Python, PyTorch, Google Cloud Platform services such as Vertex AI, Cloud Storage and PostgreSQL Cloud,
6. The database of 2000 artificial images per class was created through computer experiments. Images have a resolution of 64x64 pixels.
7. Analysis was conducted to assess image quality based on IS and FID metrics. Obtained values are: IS – 3.025, FID – 68.